%% file: Yandan_ASPDAC_2017.tex
\documentclass[twocolumn]{article}
\usepackage{graphicx}
\usepackage{amsmath,bm}
\usepackage{subfigure}
\usepackage{subfloat}

\advance\topmargin-1in\advance\oddsidemargin-1in
\evensidemargin\oddsidemargin

\makeatletter
\def\@normalsize{\@setsize\normalsize{12pt}\xpt\@xpt
\abovedisplayskip 10pt plus2pt minus5pt\belowdisplayskip \abovedisplayskip
\abovedisplayshortskip \z@ plus3pt\belowdisplayshortskip 6pt plus3pt
minus3pt\let\@listi\@listI}

\def\section{\@startsection {section}{1}{\z@}{20pt plus 2pt minus 2pt}
{8pt plus 2pt minus 2pt}{\centering\normalsize\sc
\edef\@svsec{\thesection.\ }}}
\def\thesection{\Roman{section}}

\def\subsection{\@startsection {subsection}{2}{\z@}{16pt plus 2pt minus 2pt}
{6pt plus 2pt minus 2pt}{\normalsize\sl
\edef\@svsec{\thesubsection.\ }}}
\def\thesubsection{\Alph{subsection}}

\long\def\@makecaption#1#2{
\vskip10pt\begin{center} #1 #2 \end{center}\par\vskip 1pt}
\def\fnum@figure{\raggedright{\footnotesize Fig. \thefigure }.%
\footnotesize}
\def\fnum@table{\footnotesize TABLE \thetable\\\footnotesize\sc}
\def\thetable{\Roman{table}}

\makeatother

\usepackage{comment}
\usepackage{color}
\usepackage{etoolbox}
\newbool{inccomment}
\setbool{inccomment}{true}
\newcommand{\hl}[1]{\ifbool{inccomment}{{\color{magenta}#1}}{}}
\newcommand{\ww}[1]{\ifbool{inccomment}{{\color{blue} #1}}{}}
\newcommand{\yd}[1]{\ifbool{inccomment}{{\color{red} #1}}{}}

\begin{document}
\date{}

\title{\Large\textbf {Classification Accuracy Improvement for Neuromorphic Computing Systems with One-level Precision Synapses}}


\author{
Yandan Wang, Wei Wen, Linghao Song, and Hai (Helen) Li\\
Dept. of Electrical \& Computer Engineering, University of Pittsburgh, Pittsburgh, PA, USA\\
\{yaw46,wew57,linghao.song,hal66\}@pitt.edu\\
}
\maketitle

{\small\textbf{
Abstract---Brain inspired neuromorphic computing has demonstrated remarkable advantages over traditional von Neumann architecture for its high energy efficiency and parallel data processing.
However, the limited resolution of synaptic weights degrades system accuracy and thus impedes the use of neuromorphic systems.
In this work, we propose three orthogonal methods to learn synapses with one-level precision, namely, distribution-aware quantization, quantization regularization and bias tuning, to make image classification accuracy comparable to the state-of-the-art. 
Experiments on both multi-layer perception and convolutional neural networks show that the accuracy drop can be well controlled within 0.19\% (5.53\%) for MNIST (CIFAR-10) database, compared to an ideal system without quantization. 
}}

\input{Introduction}

\input{Preliminary}
\input{Methodology}

\input{Experiments}

\input{Conclusions}

\vspace{-9pt}
\section*{Acknowledgment}
\vspace{-3pt}
This work was supported in part by NSF CCF-1615475, NSF XPS-1337198 and AFRL FA8750-15-2-0048.
Any opinions, findings and conclusions or recommendations expressed in this material are those of the authors and do not necessarily reflect the views of grant agencies or their contractors.

\end{document}

%% file: Introduction.tex
\vspace{-6pt}
\section{Introduction}
\vspace{-3pt}
In recent years, brain-inspired neuromorphic computing systems have been extensively studied. 
For example, IBM \textit{TrueNorth} 
has demonstrated many important features including high computing efficiency, extremely low power consumption, and compact volume~\cite{cassidy2013cognitive}. 
Integrating emerging technologies potentially enables a more compact and energy-efficient platform for information processing~\cite{hu2014memristor}. 
For instance, the two-terminal nonlinear memristor presents a series of advantages of good scalability, high endurance and ultra-low power consumption~\cite{gaba2014memristive}. 
Thus it is taken as a promising candidate for neuromorphic computing system development.


Neuromorphic hardware implementations usually face a major challenge on system accuracy.
TrueNorth, for example, allows only a few synaptic weights (\textit{e.g.}, $0,\pm1,\pm2$). 
Accuracy degradation is inevitable when directly deploying a learned model to the system with limited precision~\cite{cassidy2013cognitive}. 
The situation remains in memristor (or RRAM) based design. 
Theoretically, nanoscale memristor can obtain continuously analog resistance.
While, a real device often can achieve only several stable resistance states 
\cite{hu2014stochastic}. 
The distinction between theoretical and actual properties results in significant accuracy loss. 


Extensive studies on learning low-resolution synapses have been performed to improve the accuracy of neuromorphic systems. 
Wen \textit{et al.} presented a new learning method for IBM TrueNorth platform which biases the learned connection probability to binary states (0/1) to hinder accuracy loss~\cite{wen2016new}. 
Neural networks with binary resolution are more suitable for generic platforms~\cite{DBLP:journals/corr/RastegariORF16}\cite{DBLP:journals/corr/CourbariauxBD15}\cite{DBLP:journals/corr/CourbariauxB16}.
BinaryConnect~\cite{DBLP:journals/corr/CourbariauxBD15} as an example can achieve comparable accuracy in deep neural networks. 
However, neither TrueNorth nor BinaryConnect are pure binary neural networks: 
TrueNorth relies on the ensemble averaging layer in floating-point precision while the last layer of BinaryConnect is a floating-point L2-SVM. 

In this work, we focus on the pure binary (1-level precision\footnote{We define \textit{n}-level quantization as quantifying each weight to the nearest value of zero, \textit{n} positive and \textit{n} negative values with the same absolute values.}) neural networks.
While the realization of continuous analogue resistance states is still challenging, the 1-level precision is well supported by most of memory materials and architectures. 
Three orthogonal methods of leaning 1-level precision synapses and tuning bias to improve image classification accuracy are proposed:

\vspace{-6pt}
\begin{itemize}
\item \textit{Distribution-aware quantization} discretizes weights in different layers to different values.
The method is proposed based on the observation that the weight distributions of a network by layers.
\vspace{-6pt}
\item \textit{Quantization regularization} directly learns a network with discrete weights during training process. 
The regularization can reduce the distance between a weight and its nearest quantization level with a constant gradient. 
\vspace{-6pt}	
\item \textit{Bias tuning} dynamically learns the best bias compensation to minimize the impact of quantization. 
It can also alleviate the impact of synaptic variation in memristor based neuromorphic systems.
\end{itemize}

%% file: Preliminary.tex
\vspace{-6pt}
\section{Preliminary}
\label{sec:prelim}
\vspace{-3pt}
\subsection{Neural Network Models}
\label{sec:prelim:NN_model}
\vspace{-3pt}
\textit{Neural networks} (NNs) are a series of models inspired by biological neuron networks. 
The function can be formulated as:
\vspace{-3pt}
\begin{equation}
\small
	\bm{y}=\bm{W}\cdot\bm{x}+\bm{b}\mathrm{~and~}
	\bm{z}=h(\bm{y}),
	\vspace{-3pt}
	\label{sec:prelim:activation_function}
\end{equation}
where the output neuron vector $\bm{z}$ is determined by the input neuron vector $\bm{x}$, the weight matrix of connections $\bm{W}$ and the bias vector $\bm{b}$. 
Usually, $h(\cdot)$ is a non-linear activation function and all the data in (\ref{sec:prelim:activation_function}) are in floating-point precision.  

\begin{figure}[t] 
		\centering
		\includegraphics[width=0.75\columnwidth]{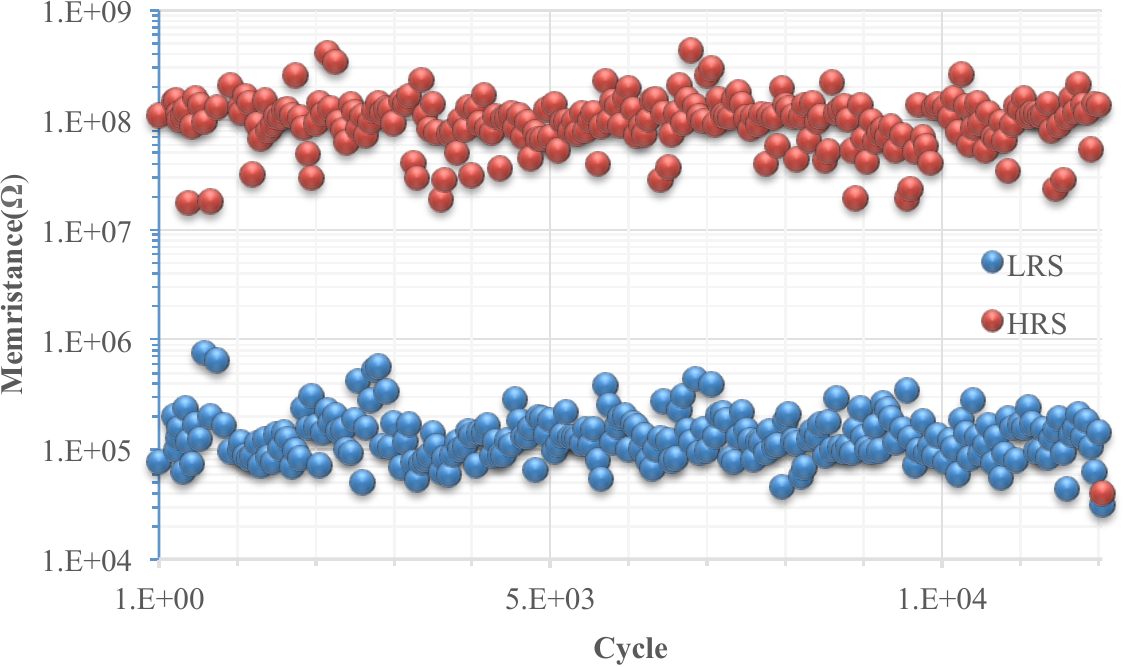}
		\vspace{-6pt}
		\caption{Statistical memristance distributions of a $TiO_2$ device.}
		\label{TiO2_memristor_value}
	\vspace{-21pt}	
\end{figure} 

\vspace{-9pt}
\subsection{Memristor Technology}
\label{sec:prelim:memristor} 
\vspace{-3pt}

Memristor, firstly introduced by Professor Leon Chua in 1971, is regarded as the fourth fundamental circuit element, representing the dynamic relationship between the charge $q(t)$ and the flux $\varphi(t)$ \cite{chua1971memristor}. 
Most significantly, the total electric flux flowing through a memristor device can be ``remembered'' by recording it as its \textit{memristance} ($M$). 
In 2008, HP Lab demonstrated the first actual memristor through a $TiO_2$ thin-film device and realized the memristive property by moving its doping front \cite{strukov2008missing}. 

Theoretically, a memristor device can achieve continuous analog resistance states. 
However, the imperfection of fabrication process causes variations and therefore memristance varies from device to device. 
Even worse, the memristance of a single memristor changes from time to time~\cite{yi2011feedback}. 
In most system designs, only two stable resistance states, \textit{high-} and \textit{low-resistance state} (HRS and LRS), are adopted. 
As the real statistical measurement data of a $TiO_2$ memristor in Fig.~\ref{TiO2_memristor_value} shows, the distribution of HRS (LRS) follows an approximated lognormal \textit{probability density function} (PDF) \cite{hu2014stochastic}.

\vspace{-9pt}
\subsection{Neuromorphic Computing Systems}
\vspace{-3pt}

\begin{figure}[t]
	\begin{center}	
		\includegraphics[width=0.65\columnwidth]{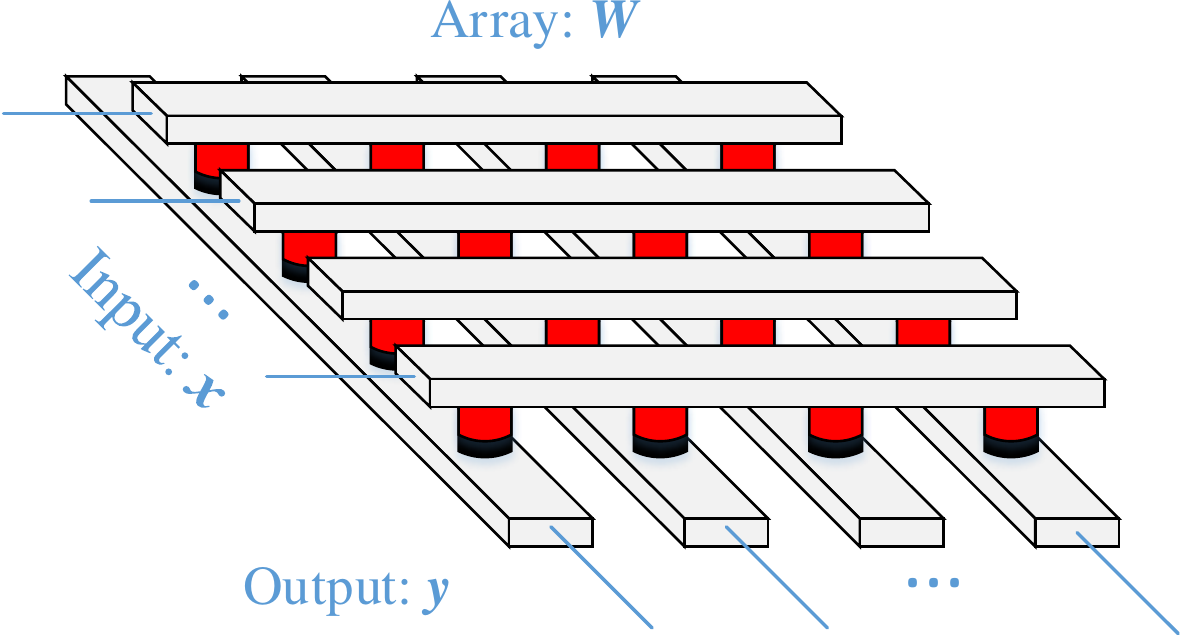}
		\vspace{-3pt}
		\caption{Mapping neural networks to memristor crossbar array.}
		\label{xbar}
	\end{center}	
	\vspace{-30pt}
\end{figure}

\textit{Neuromorphic computing systems} (NCS) represents the hardware implementations of NNs by mimicking the neuro-biological architectures. 
For example, IBM TrueNorth chip is made of a network of neuro-synaptic cores, each of which includes a 
configurable synaptic crossbar connecting 256 axons and 256 neurons in close proximity 
\cite{cassidy2013cognitive}.
The synaptic weight in the crossbar can be selected from 4 possible integers.
Memristor based NCS has also be investigated~\cite{hu2012hardware}.
Matrix-vector multiplication, the key operation in NNs, can be realized by memristor crossbar arrays as illustrated in Fig.~\ref{xbar}~\cite{wen2015eda}. 
The conductance matrix of memristor crossbar array is utilized as the weight matrix of NNs \cite{hu2012hardware}. 

The synaptic weights in these neuromorphic computing systems usually have a limited precision, constrained either by design cost (\textit{e.g.}, the SRAM cells for each weight representation in TrueNorth) or current technology process (\textit{e.g.}, two or only a few resistance levels of memristor devices).
As such, the classification accuracy loss could be very significant in NCS. 
To improve the classification accuracy, lots of research has been done \cite{DBLP:journals/corr/CourbariauxBD15}\cite{DBLP:journals/corr/CourbariauxB16}\cite{kim2016bitwise}. 
Even though, some of them have floating-point layers and some ignore circuit design constraints. 
In this work, we focus on pure binary neural networks considering the constraints in NCS hardware implementation.

%% file: Methodology.tex
\vspace{-9pt}
\section{Methodology}
\label{sec:metho}
\vspace{-6pt}

This paper aims at improving the classification accuracy of pure binary neural networks in all layers.
Such neural networks can be naturally implemented on NCS, such as TrueNorth chip and memristor based design. 
Three novel classification accuracy improving methods are proposed in the work, namely, distribution-aware quantization, quantization regularization and bias tuning. 
The implementation of \textit{convolutional neural network} (CNN) convolution operation in memristor crossbar array and a crossbar variation demo for accuracy improvement are also presented.

To explain our methodologies, in this section, we take LeNet \cite{lecun1998gradient} as the example of CNN trained on MNIST -- a 28$\times$28 handwritten digit database \cite{lecun1998mnist}. 
Experiments and analysis on more neural networks and databases shall be presented in Section~\ref{sec:experiments}.

\input{method_quantization}
\input{method_regularization}
\input{method_bias}
\input{method_cnn}

%% file: method_quantization.tex
\vspace{-9pt}
\subsection{Distribution-aware Quantization}
\label{sec:metho:quan} 
\vspace{-3pt}

\begin{figure}[t] 
	\begin{center}
		\includegraphics[width=0.95\columnwidth]{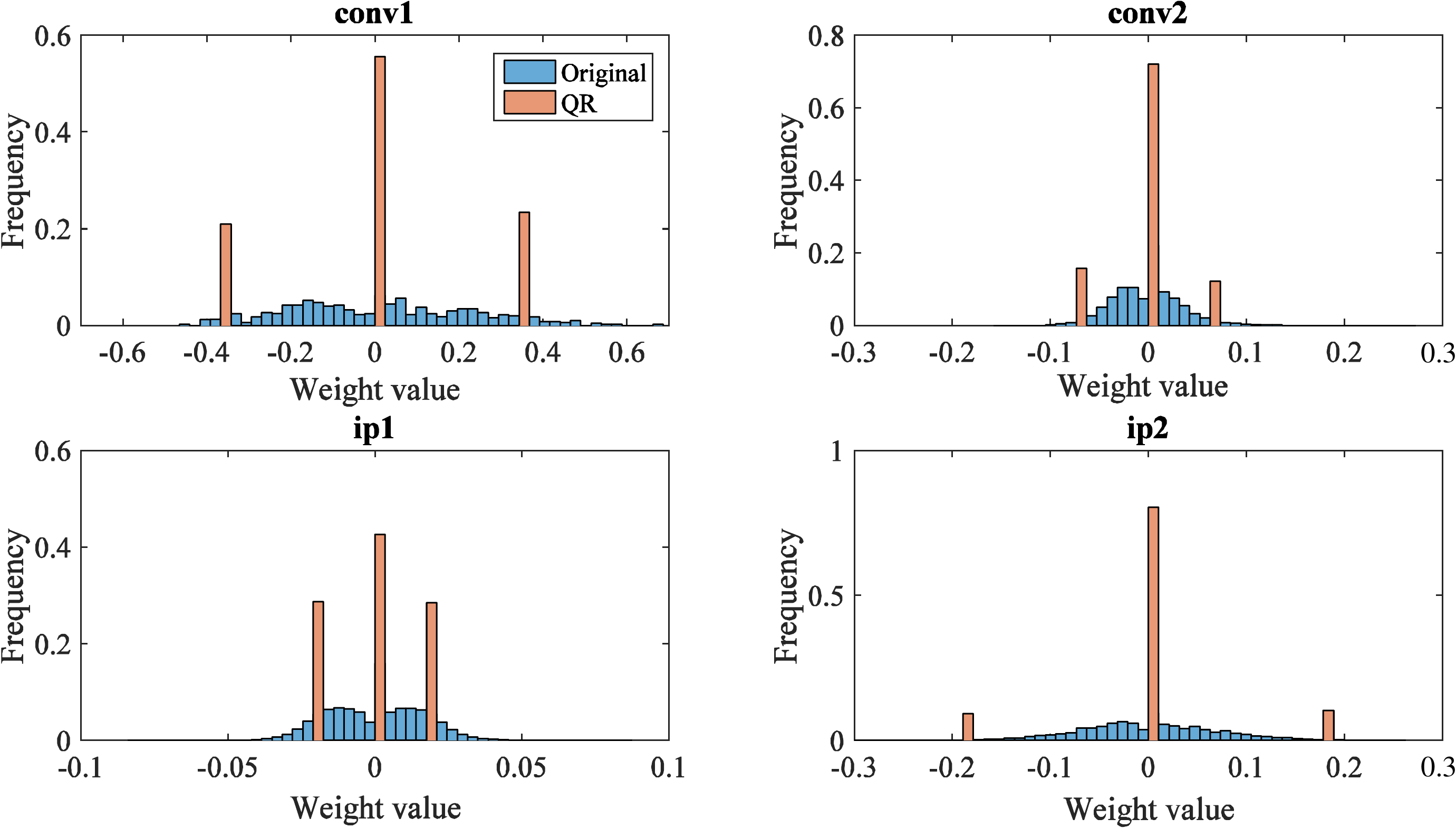}		
		\vspace{-6pt}
		\caption{The blue and orange bars denote the original weight distribution of different layers and the learned discrete weights after quantization regularization (QR) in LeNet, respectively.} 
		\label{MNIST_CNN_weight_distribution}
		\vspace{-30pt}
	\end{center}
\end{figure}

In training of neural networks, $\ell_2$-norm regularization is commonly adopted to avoid over-fitting. 
With $\ell_2$-norm regularization, the final distribution of learned weights in a layer approximately follows the normal distribution \cite{glorot2010understanding}. 
A naive quantization method in implementation is to quantify all weights to the same group of level selection. 
However, as shown in Fig.~\ref{MNIST_CNN_weight_distribution} (blue bars) by taking LeNet as an example, the weight distribution varies from layer to layer: 
The first convolutional layer (conv1) has the most scattered distribution with a wider range scope, while the weights of second convolutional layer (conv2) and two fully connected layers (ip1, ip2) have concentrated to a relatively narrow scope. 
The data implies that a quantization optimized for one layer may result in a large information loss in another layer. 

Here, we propose a heuristic method -- \textit{distribution-aware quantization} (DQ) which discretizes weights in different layers to different values. 
In memristor-based NCS, this can be realized by programming the resistance states of each crossbar to different values \cite{hu2012hardware}.
Our experiments on LeNet show that when applying the aforementioned naive method, the test accuracy of 1-level quantization quickly drops from 99.15\% to 90.77\%, while our proposed distribution-aware quantization can still achieve 98.31\% accuracy. 
Note that without explicit mention, the quantization levels are selected by cross-validation \cite{golub1979generalized}.

%% file: method_regularization.tex
\vspace{-9pt}
\subsection{Quantization Regularization}
\label{sec:metho:regularization} 
\vspace{-3pt}

Distribution-aware quantization separates the training and quantifying processes and therefore it cannot avoid the accuracy loss once the quantization is completed. 
To further improve system performance, we propose \textit{quantization regularization} (QR) which directly learns a neural network with discrete weights.

During the training of a network, a regularization term can be added to the error function to control the distribution of weights and avoid overfitting	. 
For example, $\ell_2$-norm regularization can learn weights with normal distribution and $\ell_1$-norm is commonly utilized 
to learn sparse networks \cite{glorot2010understanding}.
The total error function to be minimized with a generic regularization term can be formulated as
\vspace{-3pt}
\begin{equation}
\small
E(\bm{W})=E_D(\bm{W})+\lambda \cdot E_W(\bm{W}),
\vspace{-3pt}
\label{eq:error_function}	
\end{equation} 
where $\lambda$ is the coefficient controlling the importance between data-dependent error $E_D(\bm{W})$ and regularization term $E_W(\bm{W})$.
$\bm{W}$ is the set of all weights in neural networks. 
%
We propose a new quantization regularization as
\vspace{-3pt}
\begin{equation}
\small
E_{W}^{q}(\bm{W})=sgn\left( W_{k}-Q(W_{k}) \right) \cdot \left( W_{k} - Q(W_{k}) \right),
\label{eq:q_norm}	
\end{equation}
where $W_k$ is the \textit{k}-th weight, 
$Q(W_{k})$ is the quantization value nearest to $W_{k}$ and $sgn(\cdot)$ is the sign function. 
After forwarding and back propagation, the weight updating with learning rate $\eta$ can be formulated as:
\vspace{-3pt}
\begin{equation}
\small
W_{k} \leftarrow W_{k} - \eta \cdot \frac{\partial {E_D(\bm{W})} }{\partial {W_{k}} } - \eta \cdot sgn( W_{k}-Q(W_{k})).
\label{eq:weight_updating}	
\end{equation}
Through the third term on the right side of~(\ref{eq:weight_updating}), our regularization descents (reduces) the distance between a weight and its nearest quantization level with a constant gradient ($\pm1$). 
Compared with the $\ell_1$-norm and $\ell_2$-norm regularization, our proposed regularization method can quantify learning weights to the desired discrete values more precisely, meanwhile properly control the weight distribution and overfitting. 

\begin{figure}[t] 
	\begin{center}
		\includegraphics[width=0.65\columnwidth]{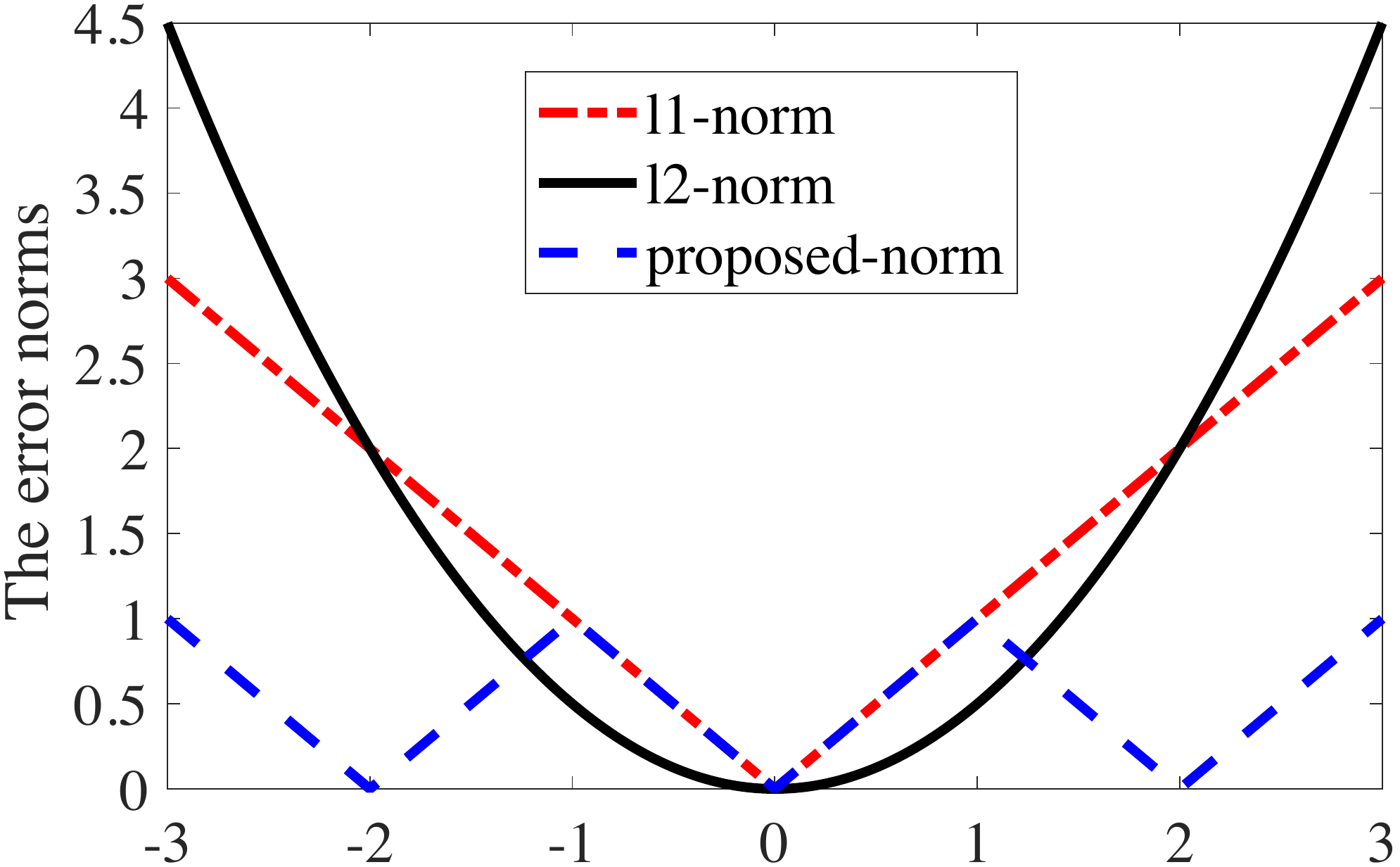}		
		\vspace{-6pt}
		\caption{Comparison of $\ell_1$-norm, $\ell_2$-norm and our proposed regularization.}
		\label{L1_L2_proposed_norm}
		\vspace{-30pt}
	\end{center}
\end{figure}

Fig.~\ref{L1_L2_proposed_norm} demonstrates and compares the three regularization methods.
Zero is one of the targeted quantification values in this work, which is usually realized through $\ell_1$-norm based neural network sparsification. 
In addition, our proposed method include more discrete quantification values.  
Orange bars in Fig.~\ref{MNIST_CNN_weight_distribution} correspond to the new weight distribution of LeNet after applying QR, indicating our method can efficiently learn weights around quantization levels.
Compared with the naive 1-level quantization, including QR only can improve accuracy 6.21\%. 
Combining with DQ, the accuracy drop from the ideal case is controlled within only 0.20\% with 1-level quantization. 
More experiments will be discussed in section~\ref{sec:experiments}.

%% file: method_bias.tex
\vspace{-9pt}
\subsection{Bias Tuning}
\label{sec:metho:bias}
\vspace{-3pt}

The quantization of weights deviating the information can be formulated as
\vspace{-3pt}
\begin{equation}
\small
	y_j + \Delta y_j = \sum_i (W_{ji} + \Delta W_{ji}) \cdot  x_{i} + b_j,
	\vspace{-6pt}
\end{equation}
where $W_{ji}$ is the weight connecting the \textit{i}-th neuron in the previous layer to the \textit{j}-th neuron in this layer.
$\Delta W_{ji}$ and $\Delta y_j = \sum_i \Delta W_{ji} \cdot x_{i}$ are the deviation of weight and input of activation function, respectively, resulted from quantization. 
The deviation $\Delta y_j$ propagates through layers toward the output classifier neurons and deteriorates the classification accuracy.

In circuit design of neuron model, the bias usually is an adjustable parameter, \textit{e.g.} the fire threshold in TrueNorth neuron model works as bias. Therefore, to compensate the deviation, we may adjust the neuron bias from $b_j$ to $b_j + \Delta b_j$ such that
\vspace{-3pt}
\begin{equation}
\small
\Delta b_j = -\Delta y_j = - \sum_i \Delta W_{ji} \cdot x_{i}.
\vspace{-6pt}
\end{equation}
As such, the neuron activation can remain the original value before quantization. 
Unfortunately, the input $x_{i}$ varies randomly with the input samples (\textit{e.g.}, images) and a unique bias compensation $\Delta b_j$ cannot be identified. 

\begin{figure}[t] 
	\begin{center}
		\includegraphics[width=0.85\columnwidth]{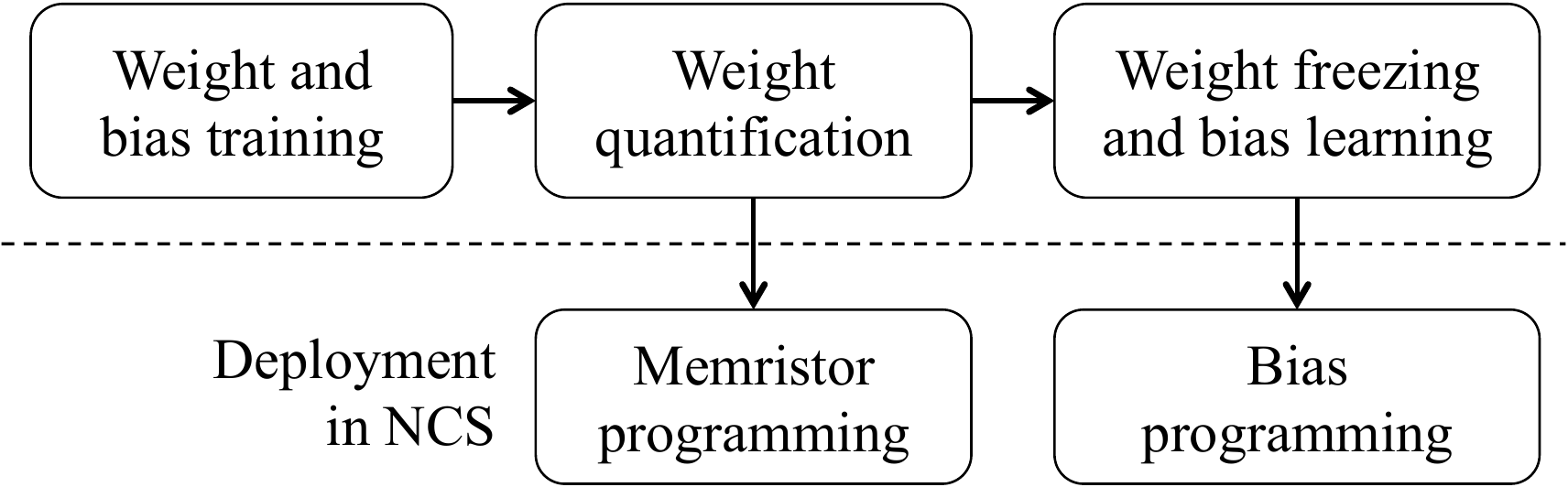}		
		\vspace{-6pt}
		\caption{The framework of proposed bias tuning method.}
		\vspace{-30pt}
		\label{biastuning}
	\end{center}
\end{figure}

We propose \textit{bias tuning} (BT) which learns the optimal bias compensation to minimize the impact of quantization. 
Fig.~\ref{biastuning} shows the framework of the bias tuning: 
first, both weights and biases are trained without quantization; 
second, weights are quantified and programmed into NCS; 
third, weights are frozen and biases are learned to improve classification accuray; 
and finally, the tuned biases are programmed into NCS. 
Impressively, bias tuning method can achieve 7.89\% classification improvement compared to the naive 1-level quantization baseline on LeNet. 
Combining with the above DQ and QR methods, the total accuracy drop can be reduced to merely 0.19\%. 

%% file: method_cnn.tex
\vspace{-9pt}
\subsection{Convolution in Memristor Crossbar Array}
\label{sec:metho:conv} 
\vspace{-3pt}

The memristor crossbar structure can be naturally mapped to fully connected layers. 
Here, we extend its use to convolution layers. 
A pixel value ($y$) in a post feature map is computed by
\vspace{-3pt}
\begin{equation} 
\label{eq:conv}
\small
y = \sum_k F_{k}\cdot w_k + b,
\vspace{-3pt}
\end{equation}
where $w_k$ is the \textit{k}-th weight in the filter and $F_{k}$ is the corresponding input feature. 
Because the essence of convolution is multiplication-accumulation, we can employ memristor crossbar array to compute. 

\begin{figure}[b] 
\vspace{-12pt}
\begin{center}
\includegraphics[width=0.85\columnwidth]{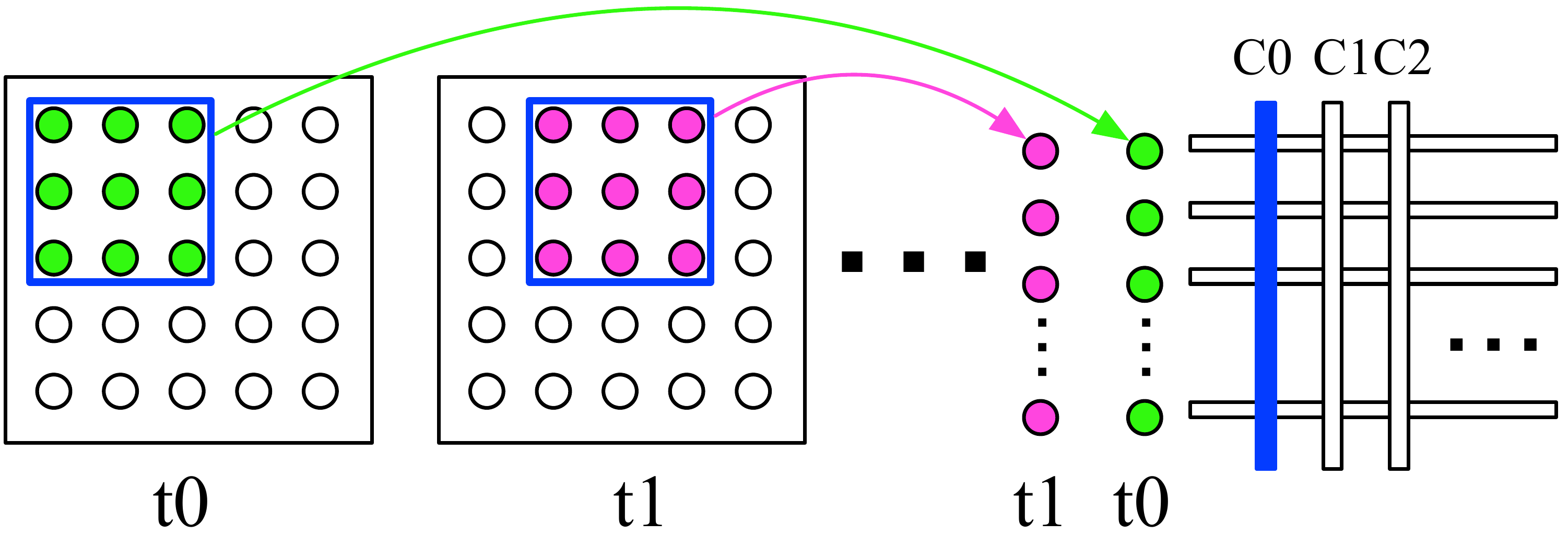}		
\vspace{-6pt}
\caption{Convolution implementation in memristor crossbar array.}
\label{Conv_ReRAM}
\end{center}
\end{figure}

Fig. \ref{Conv_ReRAM} shows an example to compute the convolution of a 5-by-5 feature map with a 3-by-3 filter. 
At the time stamp \textit{t0}, the green elements are converted to a vector and sent into a memristor array through word lines. 
And at \textit{t1}, the pink elements are processed similarly to the green ones. 
As the filter shifts, the corresponding features in the previous layer are sent into the crossbar in a time-division sequence, such that the output features are computed by the bit line (blue) whose weights belong to the filter.
As shown in the figure, each bitline is mapped to one filter in the convolutional layer.
We note that the proposed DQ, DR and BT methods also work for weights in CNN.

%% file: Experiments.tex
\vspace{-9pt}
\section{Experiments}
\label{sec:experiments}
\vspace{-3pt}

\input{experiment_setup}

\input{experiment_mlp_mnist}
\input{experiment_cnn_mnist}

\input{experiment_cnn_cifar10}
\input{experiment_filters}
\input{experiment_variance}
\input{experiment_evaluation}

%% file: experiment_setup.tex
\subsection{Experiment Setup}
\label{sec:experiment:setup}
\vspace{-3pt}

\input{tables/experiment_setup}

To evaluate the effectiveness of proposed methods, we conducted three experiments using \textit{multilayer perception} (MLP) and CNN neural network structures on two datasets: MNIST and CIFAR-10 (a 32$\times$32 color image database). 
The first two experiments are both conducted on MNIST dataset using a MLP and a CNN network, respectively. 
The third experiment is conducted on CIFAR-10 dataset using a CNN network. 
The adopted deep leaning framework is Caffe developed by the \textit{Berkeley Vision and Learning Center} (BVLC) and community contributors \cite{jia2014caffe}. 
Detailed network parameters and dataset are summarized in Table~\ref{tab:fc:setup}.

%% file: tables/experiment_setup.tex
\begin{table}[t]
	\caption{Network and dataset}
	\label{tab:fc:setup}
	\vspace{-6pt}
	\centering
	\footnotesize
		\def\arraystretch{1}\tabcolsep 8.5pt
		\begin{tabular}{c c c c}			
			\hline
			\parbox[c]{1mm}{} & Network 1 & Network 2 & Network 3\\
			\hline
			Dataset   & MNIST           & MNIST                     & CIFAR-10\\
			Input     & 28$\times$28    & 28$\times$28              & 32$\times$32\\
			Conv1     & $-$             & 20$\times$5$\times$5\textsuperscript{\textsection}      & 32$\times$5$\times$5\\   
			Conv2     & $-$             & 50$\times$5$\times$5      & 32$\times$5$\times$5\\ 
		    Conv3     & $-$             & $-$                       & 64$\times$5$\times$5\\   
			Ip1       & 784$\times$500  & 800$\times$500               & 1024$\times$10\\
			Ip2       & 500$\times$300  & 500$\times$10                & $-$\\
			Ip3       & 300$\times$10   & $-$                       & $-$\\   
			\hline
		\end{tabular}
		
		\vspace{3pt}
		\textsuperscript{\textsection}20$\times$5$\times$5 means 20 filters with each filter size 5$\times$5.
	\vspace{-12pt}
\end{table}

%% file: experiment_mlp_mnist.tex
\vspace{-9pt}
\subsection{Function Validation of MLP on MNIST}
\label{sec:experiment:MLP}
\vspace{-3pt}

Network 1 is a MLP network with a size of $784\times500\times300\times10$, which can't be directly implemented in NCS. 
Previously, we presented the hardware implementation of mapping a large network to small crossbar arrays \cite{wen2015eda}. 
Here, 784 corresponds to the 28$\times$28 MNIST image input pattern; 500 and 300 are the neuron numbers of the first and second hidden layers, respectively; and 10 is the final classification outputs. 

The baseline is set as the highest accuracy (all the layers quantified to 0.06) of all naive 1-level quantization situations without applying any proposed method. 
To explore the effectiveness of each single method and their combination situations, we conducted 8 separate experiments with combinations, the experiment results of which are summarized in Table~\ref{tab:fc:mlp_mnist}.     

\input{tables/mlp_mnist}

Compared with the baseline accuracy, there is a large accuracy increase when applied only one of three accuracy improvement methods (1.52\%, 1.26\%, 0.4\%, respectively). 
Applying any two of three methods will make the accuracy further increased. 
Combining all three methods together can achieve a highest accuracy with only 0.39\% accuracy drop compared with the ideal value without any quantization. 
We note that, in some cases (\textit{e.g.} DQ+QR+BT vs. DQ+BT), integrating more than one proposed methods does not improve accuracy much.
This is because MNIST is a relative simpler database so the effectiveness of these methods on accuracy improvement quickly approaches to a saturated level. 
In more challenging CIFAR-10 database, experiments show that more methods of DQ, QR and BT are harnessed, higher accuracy can always be obtained by a large margin. 

%% file: tables/mlp_mnist.tex
\begin{table}[b]
	\vspace{-18pt}
	\caption{The accuracy measurement for MLP on MNIST dataset}
	\label{tab:fc:mlp_mnist}
	\vspace{-5pt}
	\centering
	\footnotesize
		\def\arraystretch{1}\tabcolsep 7pt
		\begin{tabular}{c c c c c c}			
			\hline
			\parbox[c]{1mm}{} & DQ & QR & BT & Accuracy & Drop\\
			\hline
			Ideal\textsuperscript{\textsection}     &  &  &   & 98.39\% &\\
			0 (Baseline)    &  &  &  & 95.97\% & 2.42\%\\	
			\hline
			  1 &         $\surd$\textsuperscript{$\ast$} & & & 97.49\% & 0.90\%\\
			  2 &         & $\surd$ & & 97.23\% & 1.16\%\\
			  3 &         & & $\surd$ & 96.37\% & 2.02\%\\
			  \hline
			  4 &         $\surd$ & $\surd$ & & 97.91\% & 0.48\%\\
			  5 &         $\surd$ & & $\surd$ & 98.00\% & 0.39\%\\
			  6 &         & $\surd$ & $\surd$  & 97.23\% & 1.16\%\\
			  \hline
			  7 &         $\surd$ & $\surd$ & $\surd$ & 98.00\% & 0.39\%\\
			\hline
		\end{tabular}
		\footnotetext{\textsuperscript{\textsection}The ideal accuracy without quantization; \textsuperscript{$\ast$}$\surd$ denotes that the corresponding method is utilized.}
		
		\vspace{3pt}
		\textsuperscript{\textsection}The ideal accuracy without quantization; 
		
		\textsuperscript{$\ast$}$\surd$ denotes that the corresponding method is utilized.
\end{table}

%% file: experiment_cnn_mnist.tex
\input{tables/cnn_mnist}

\vspace{-9pt}
\subsection{Function Validation of LeNet}
\label{sec:experiment:CNN_MNIST}
\vspace{-3pt}

LeNet, which has strong robustness to image geometric transformations, is a much more popular network.
We utilized it for MNIST and shows the results in Table~\ref{tab:fc:cnn_mnist}. 
Compared with the MLP network, 1-level precision LeNet can achieve an even lower accuracy drop (0.19\% compared with 0.39\%) after combining all our methods. 
Remarkably, although the DQ method separates the training and quantifying processes, directly quantifying weights in each layer has accuracy loss less than 1\%, without further fine-tuning. 
The orthogonality among DQ, QR and BT is also indicated by the results.

%% file: tables/cnn_mnist.tex
\begin{table}[t]
	\caption{The accuracy measurement for CNN on MNIST dataset}
	\label{tab:fc:cnn_mnist}
	\vspace{-5pt}
	\centering
	\footnotesize
		\def\arraystretch{1}\tabcolsep 7pt
		\begin{tabular}{c c c c c c}			
			\hline
			\parbox[c]{1mm}{} & DQ & QR & BT & Accuracy & Drop\\
			\hline
			Ideal       &  &  &   & 99.15\% &\\
			0 (Baseline)    &  &  &   & 90.77\% & 8.38\%\\	
			\hline
			1 &         $\surd$ & & & 98.31\% & 0.84\%\\
			2 &         & $\surd$ & & 96.98\% & 2.17\%\\
			3 &         & & $\surd$ & 98.66\% & 0.49\%\\
			  \hline
			4 &         $\surd$ & $\surd$ & & 98.96\% & 0.19\%\\
			5 &         $\surd$ & & $\surd$ & 98.68\% & 0.47\%\\
			6 &         & $\surd$ & $\surd$  & 98.75\% & 0.40\%\\
			  \hline
			7 &         $\surd$ & $\surd$ & $\surd$ & 98.96\% & 0.19\%\\
			\hline
		\end{tabular}
		\footnotetext[1]{\scriptsize Uppercase}
	\vspace{-12pt}
\end{table}

%% file: experiment_cnn_cifar10.tex
\vspace{-9pt}
\subsection{Function Validation of CNN on CIFAR-10}
\label{sec:experiment:CNN_CIFAR10}
\vspace{-3pt}

\input{tables/cnn_cifar10}

We also evaluate the proposed methods in more challenging natural image dataset CIFAR-10 to verify their generality. 
The CNN in \cite{alexnet} is adopted without data augmentation. 
Table~\ref{tab:fc:cnn_cifar10} presents the results of all the interested combinations.

As expected, CNN has a large accuracy drop (64.32\%) when applying the naive 1-level quantization while each our proposed technique can dramatically hinder the accuracy loss.
However, unlike the experiments on MNIST, a sole method cannot improve the accuracy of CNN to a satisfactory level. 
Some combinations of two methods perform excellent accuracy improvement.
For example, DQ+RQ makes the accuracy level to 74.43\%

BinaryConnect neural network in \cite{DBLP:journals/corr/CourbariauxBD15} performs state-of-the-art accuracy when the last layer utilizes L2-SVM. 
The parameters in the L2-SVM layer are floating-point and critical for accuracy maintaining. 
However, the SVM is not good for circuit implementation. 
Our work quantifies all weights to one level and controls the accuracy loss within 5.53\% for more efficient circuit (\textit{e.g.}, memristor crossbar) design. 

%% file: tables/cnn_cifar10.tex
\begin{table}[b]
	\vspace{-18pt}
	\caption{The accuracy measurement for CNN on CIFAR-10 dataset}
	\vspace{-4pt}
	\centering
	\footnotesize
	\label{tab:fc:cnn_cifar10}
		\def\arraystretch{1}\tabcolsep 7pt
		\begin{tabular}{c c c c c c}			
			\hline
			\parbox[c]{1mm}{} & DQ & QR & BT & Accuracy & Drop\\
			\hline
			Ideal       &  &  &   & 82.12\% &\\
			0 (Baseline)    &  &  &   & 17.80\% & 64.32\%\\	
			\hline
			 1 &         $\surd$ & & & 32.92\% & 49.2\%\\
			 2 &         & $\surd$ & & 66.88\% & 15.24\%\\
			 3 &         & & $\surd$ & 46.54\% & 35.58\%\\
			  \hline
			 4 &         $\surd$ & $\surd$ & & 74.43\% & 7.69\%\\
			 5 &         $\surd$ & & $\surd$ & 57.74\% & 24.38\%\\
			 6 &         & $\surd$ & $\surd$  & 67.22\% & 14.90\%\\
			  \hline
			 7 &         $\surd$ & $\surd$ & $\surd$ & 76.59\% & 5.53\%\\
			\hline
		\end{tabular}
		\footnotetext[1]{\scriptsize Uppercase}
\end{table}

%% file: experiment_filters.tex
\vspace{-9pt}
\subsection{Learned Filters}
\label{sec:experiment:filters}
\vspace{-3pt}

\begin{figure}[t]
	\begin{center}	
		\includegraphics[width=1\columnwidth]{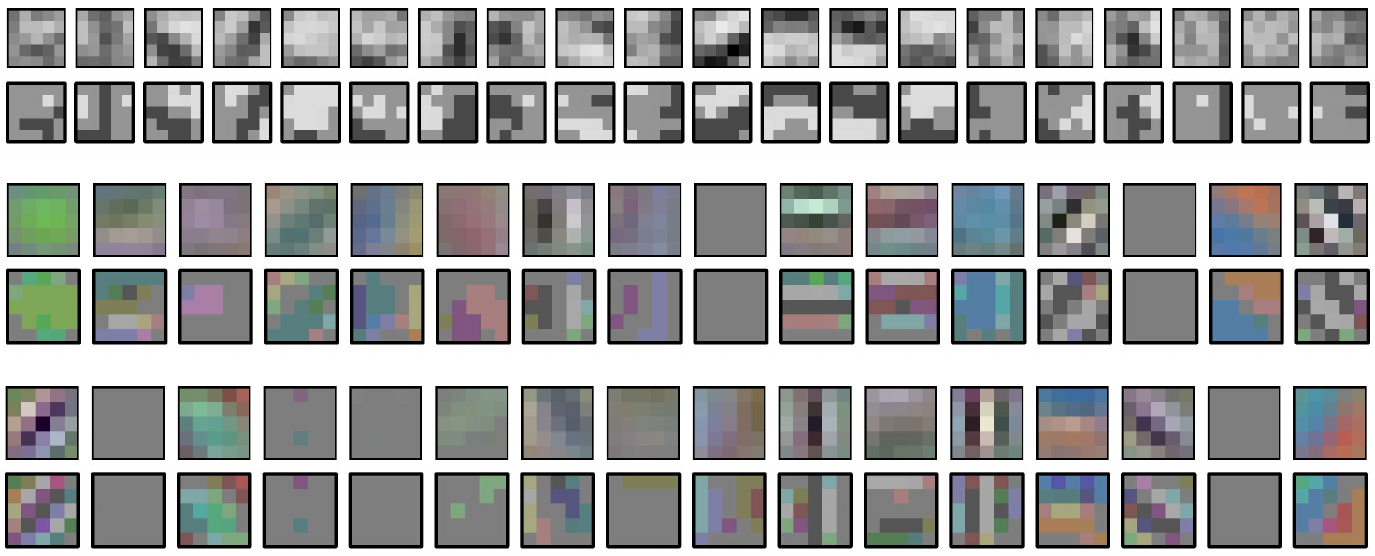}
		\vspace{-18pt}
		\caption{The learned floating-point (upper) and quantified (lower) \textit{conv1} filters in LeNet (the gray-scale ones) and CNN on CIFAR-10 (the color ones). A zero weight is mapped to pixel value 128, and negative (positive) weights are darker (brighter) ones.}
		\label{cropped_filter}
		\vspace{-30pt}
	\end{center}	
\end{figure}

Fig.~\ref{cropped_filter} presents the learned floating-point and 1-level precision \textit{conv1} filters in LeNet and CNN on CIFAR-10, respectively. 
Our methods can efficiently learn the feature extractors similar to the corresponding original ones, even with 1-level precision. 
Furthermore, the number of input channels (RGB) of CIFAR-10 image is 3, such that each pixel in the filter has $3^3$ possible colors. 
For filters with $n$ channels, a 1-level precision filter still has a large learning space with $3^{n \cdot k \cdot k}$ possibilities, where $k$ is the filter size. 
Those explain why our method can maintain the comparable accuracy.

%% file: experiment_variance.tex
\vspace{-9pt}
\subsection{Bias Tuning to Alleviate Crossbar Variation}
\label{sec:experiment:variation}
\vspace{-3pt}

As aforementioned, the memristive variations caused by fabrication imperfection can result in deviation of the programmed weights \cite{hu2014stochastic}.
Our bias tuning method can also be extended to overcome memristor variation. 
After programming weights to memristors under the impact of variation, we read out the real programmed weights, then fine-tune the bias with weights frozen, and finally the tuned biases are reprogrammed to the circuit neuron models to compensate the impact of weight variation. 

\begin{figure}[tb] 
	\begin{center}
	\includegraphics[width=0.9\columnwidth]{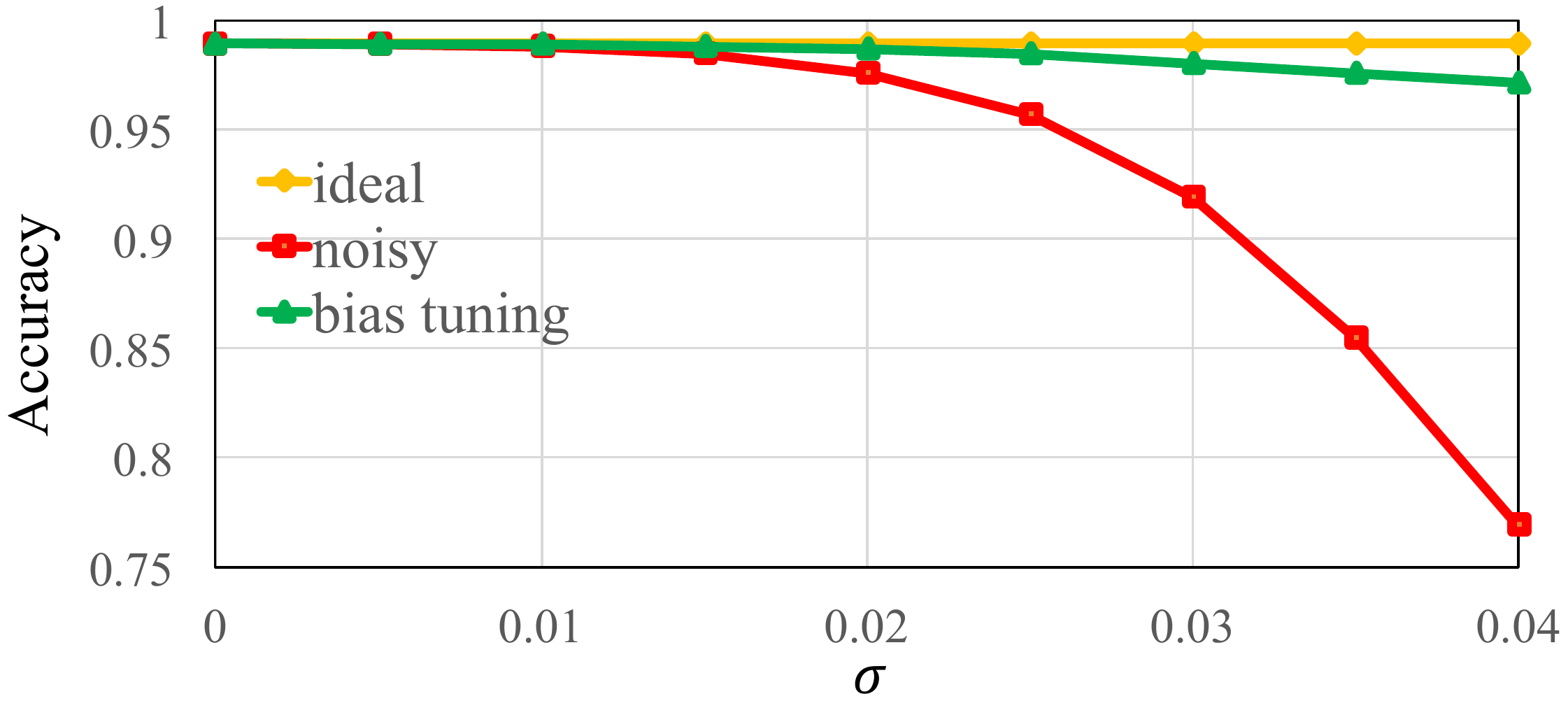}
	\vspace{-9pt}
	\caption{The bias tuning in LeNet. The yellow line denotes the accuracy after applying DQ and QR without noise; The red line is the baseline with quantization and noise; The green line denotes the accuracy recovered from the baseline after bias tuning; $\sigma$ is the standard deviation of Gaussian noise. }
	\label{noise}
	\end{center}
	\vspace{-28pt}	
\end{figure}

Fig. \ref{noise} plots the accuracy vs. the variance of programming process. 
The entry 4 in Table~\ref{tab:fc:cnn_mnist} is taken as the baseline in this investigation on variation impact. 
The figure shows that the bias tuning method successfully hinders the negative impact of variation.

%% file: experiment_evaluation.tex
\vspace{-9pt}
\subsection{Discussion}
\label{sec:experiment:evaluation}
\vspace{-3pt}

Our previous research study \cite{wen2016new} specifies for spiking neural networks, where the probability distribution can only be biased to two poles (0 or 1).
In this work, we extend the method to memristor-based neural networks adopted by state-of-the-art research and large-scale applications \cite{tang2015spiking}.
The proposed methods can regularize the floating-point weights to multiple levels with uniform or nonuniform quantization.
For example in our CIFAR-10 experiments, the quantization points in layer conv1, conv2, conv3 and ip1 are $[-0.12,0,0.12]$, $[-0.08,0,0.08]$, $[-0.02,0,0.02]$ and $[-0.008,0,0.008]$, respectively. 
Moreover, we discharge the reliance on the floating-point layer in \cite{wen2016new} and explore a pure one-level precision solution. 
Comprehensive experiments and analyses on MLP and CNN using MNIST and CIFAR-10 datasets are conducted. 
Our experiments on MNIST shows negligible accuracy drop (0.19\% in CNN), which is much better than the previous work like \cite{wen2016new}.

From the aspect of the system implementation, there are extensive research studies on binary neural networks 
deployed in traditional platforms such as CPUs, GPUs and FPGAs. 
However, those approaches may not suitable for the hardware characteristics of brain-inspired systems like memristor-based systems. 
For example, BinaryConnect \cite{DBLP:journals/corr/CourbariauxBD15} uses L2-SVM layer, which is very costly to be implemented by memristor hardware. 
In circuit design, \textit{bias} has the characteristic of adjustability, which inspires our bias tuning method in this work. 
As shown in the paper, bias tuning can be used to control quantization accuracy as well as overcome the process variation of memristor technology.

%% file: Conclusions.tex
\vspace{-9pt}
\section{Conclusions}
\label{sec:conclusions}
\vspace{-6pt}
In this work, we analyze the impact on accuracy degradation of low-resolution synapses in neuromorphic hardware implementations theoretically and propose three orthogonal methods to learn synapses with 1-level precision. 
We applied these proposed methods and their combinations to MLP on MNIST, CNN on MNIST and CNN on CIFAR-10 database, comparable state-of-the-art achievements are obtained: only 0.39\%, 0.19\%, and 5.53\% accuracy loss, respectively.
Our work will be more suitable for memristor-based neural networks.